\pdfoutput=1

\documentclass[11pt]{article}

\usepackage[preprint]{acl}

\usepackage{times}
\usepackage{latexsym}
\usepackage{stfloats} 

\usepackage[T1]{fontenc}

\usepackage[utf8]{inputenc}

\usepackage{microtype}

\usepackage{inconsolata}

\usepackage{booktabs, tabularx, multirow, multicol, makecell, colortbl}

\usepackage{graphicx}
\usepackage{xcolor}
\usepackage{amsmath}

\title{MIBench: Evaluating Multimodal Large Language Models \\ over Multiple Images}

\author{
\textbf{Haowei Liu}$^{1,2}$, \textbf{Xi Zhang}$^{3}$, \textbf{Haiyang Xu}$^{3\dagger}$,
\textbf{Yaya Shi}$^{4}$, \textbf{Chaoya Jiang}$^{5}$, \textbf{Ming Yan}$^{3}$, \\
\textbf{Ji Zhang}$^{3}$, \textbf{Fei Huang}$^{3}$, \textbf{Chunfeng Yuan}$^{1,2\dagger}$,
\textbf{Bing Li}$^{1,2}$, \textbf{Weiming Hu}$^{1,2,6}$ \\
$^{1}$MAIS, Institute of Automation, Chinese Academy of Sciences, China \\
$^{2}$School of Artificial Intelligence, University of Chinese Academy of Sciences, China \\
$^{3}$Alibaba Group
$^{4}$University of Science and Technology of China
$^{5}$Peking University \\
$^{6}$School of Information Science and Technology, ShanghaiTech University, China \\
\texttt{liuhaowei2019@ia.ac.cn}, \texttt{cfyuan@nlpr.ia.ac.cn} \\
\texttt{\{shuofeng.xhy, ym119608\}@alibaba-inc.com} \\
}

\begin{document}
\maketitle

\let\thefootnote\relax\footnotetext{$^\dagger$Corresponding authors.}

\begin{abstract}

Built on the power of LLMs, numerous multimodal large language models (MLLMs) have recently achieved remarkable performance on various vision-language tasks. However, most existing MLLMs and benchmarks primarily focus on single-image input scenarios, leaving the performance of MLLMs when handling realistic multiple images underexplored. Although a few benchmarks consider multiple images, their evaluation dimensions and samples are very limited. In this paper, we propose a new benchmark \textbf{MIBench}, to comprehensively evaluate fine-grained abilities of MLLMs in multi-image scenarios. Specifically, MIBench categorizes the multi-image abilities into three scenarios: multi-image instruction (MII), multimodal knowledge-seeking (MKS) and multimodal in-context learning (MIC), and constructs 13 tasks with a total of 13K annotated samples. During data construction, for MII and MKS, we extract correct options from manual annotations and create challenging distractors to obtain multiple-choice questions.
For MIC, to enable an in-depth evaluation, we set four sub-tasks and transform the original datasets into in-context learning formats. We evaluate several open-source and closed-source MLLMs on the proposed MIBench. The results reveal that although current models excel in single-image tasks, they exhibit significant shortcomings when faced with multi-image inputs, such as limited fine-grained perception, multi-image reasoning and in-context learning abilities. The annotated data of MIBench is available at \url{https://huggingface.co/datasets/StarBottle/MIBench}.
\end{abstract}

\section{Introduction}

\begin{figure}[t]
    \centering
    \includegraphics[width=0.8\columnwidth]{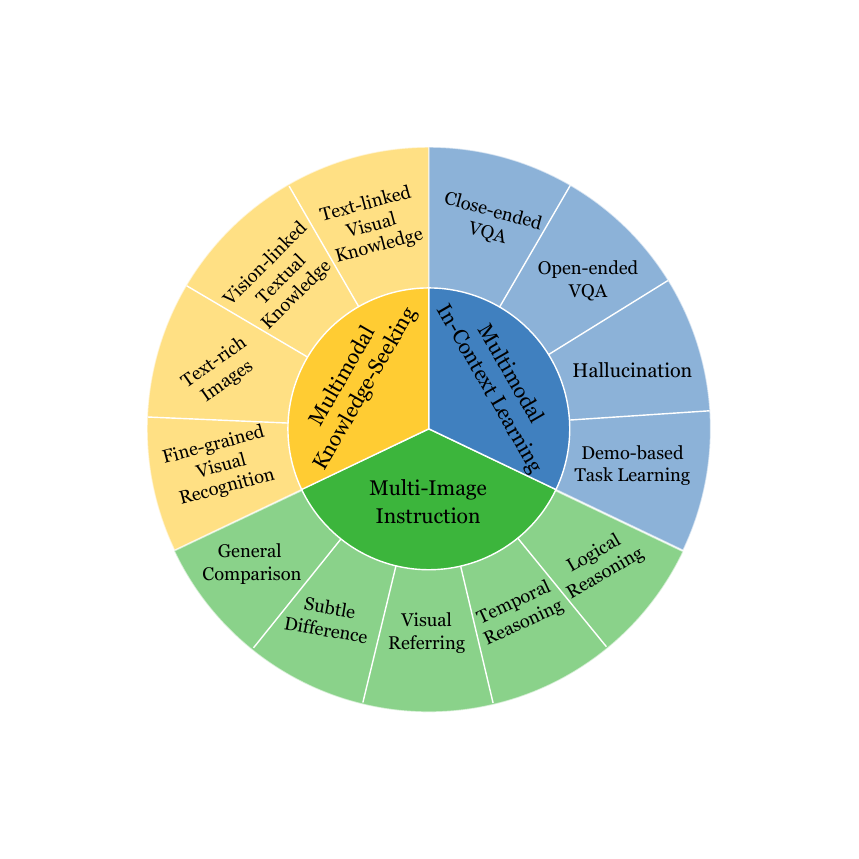}
    \caption{Overview of our MIBench, which covers three multi-image scenarios and a total of 13 tasks.}
    \label{fig: overview}
    \vspace{-2mm}
\end{figure}

\begin{table*}[t]
\centering
\small
\setlength{\tabcolsep}{2pt}
\resizebox{0.98\textwidth}{!}{
\begin{tabular}{lccccc}
\toprule
\textbf{Benchmark} & \textbf{Scenario} & \textbf{\#Multi-Image Task} & \textbf{\#Multi-Image Sample} & \textbf{Answer Type} & \textbf{Evaluator} \\
\midrule
MME~ & Single-Image & 0 & 0 & Yes/No & Metrics \\
MMBench~ & Single-Image & 0 & 0 & Multi-choice & GPT \\
SEED-Bench~ & Single-Image & 4 & 829 & Multi-choice & Metrics \\
\midrule
Sparkles-Eval~ & Multi-Image Dialogue & 1 & 150 & Open-ended & GPT-4 \\
Mantis-Eval~ & Multi-Image Reasoning & 1 & 217 & Multi-choice \& Short Answer & Metrics \\
\midrule
MIBench & Comprehensive Multi-Image & 13 & 13K & Multi-choice \& Short Answer & Metrics \\\bottomrule
\end{tabular}}
\caption{Comparison of the proposed MIBench with recent MLLM benchmarks.}
\label{tab:benchmarks}
\vspace{-2mm}
\end{table*}

Recently, leveraging the powerful comprehension and reasoning abilities of LLMs, many MLLMs such as LLaVA-1.5 \citep{liu2024visual} and mPLUG-Owl2 \citep{ye2024mplug} have demonstrated outstanding performance across various vision-language tasks (\textit{e.g.} image captioning, VQA and visual grounding). Concurrently, numerous benchmarks like MME \citep{fu2023mme}, MMBench \citep{liu2023mmbench} and SEED-Bench \citep{li2024seed}) have been proposed to evaluate the abilities of MLLMs in terms of different perspectives such as recognition, localization and reasoning.

However, most existing MLLMs focus on single-image scenarios. Accordingly, previous benchmarks primarily evaluate MLLMs based on single-image inputs.
In contrast, real-world multimedia information, such as web pages and social media, generally contains multiple images and corresponding text in interleaved forms. 
Therefore, multi-image scenarios have greater practical value than single-image scenarios, which makes it worth exploring whether existing single-image MLLMs possess emergent abilities for multi-image inputs.
Moreover, some methods like Sparkles \citep{huang2023sparkles} and Mantis \citep{jiang2024mantis} explore multi-image scenarios but have not comprehensively evaluated the models' multi-image abilities.
As shown in Table \ref{tab:benchmarks}, Sparkles evaluates the model solely on a small-scale multi-image chat dataset, and the assessment relies entirely on scoring by GPT-4.
Mantis-Eval focuses on multi-image reasoning and has a limited scale of 217 samples.

\begin{figure*}
    \centering
    \includegraphics[width=\linewidth]{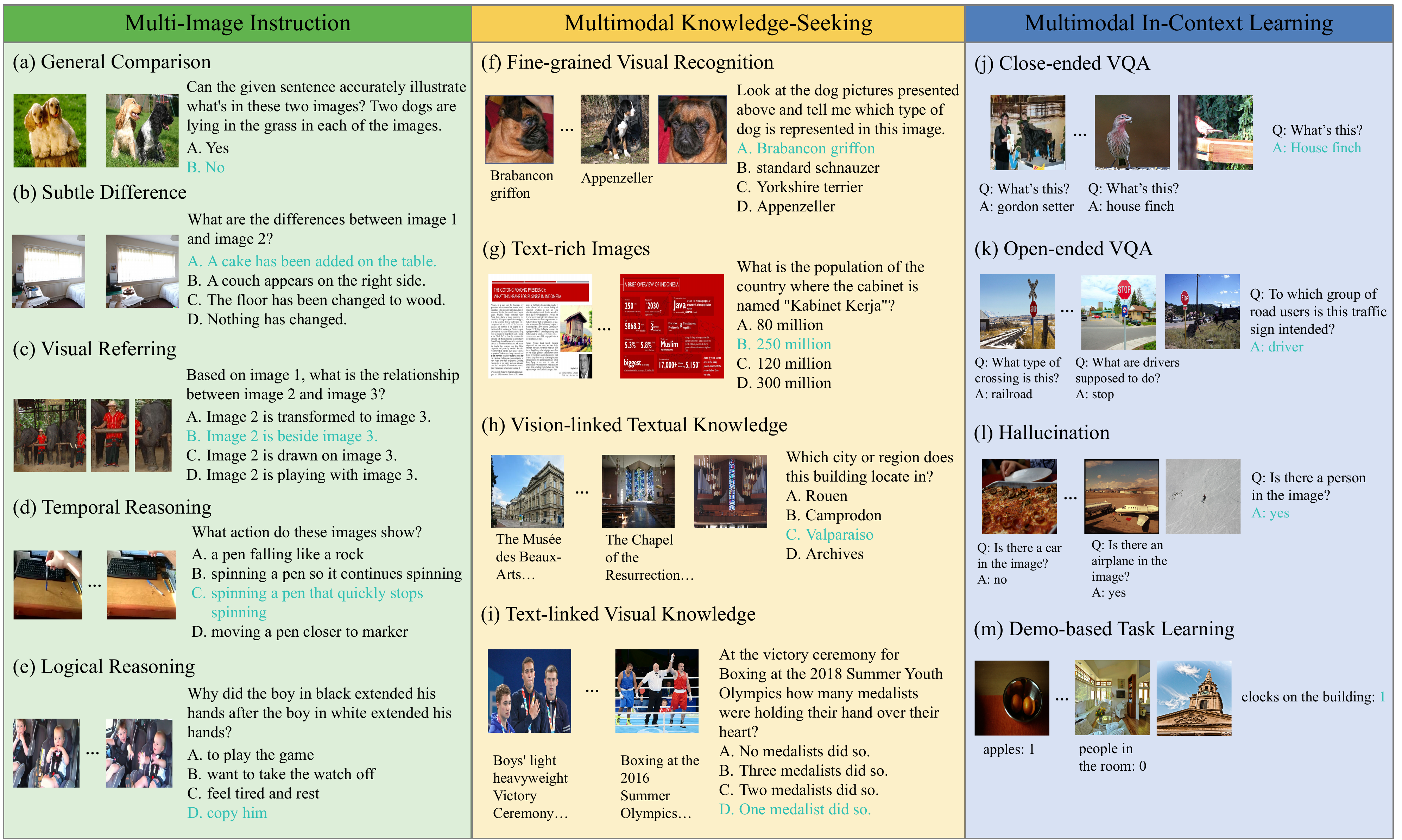}
    \caption{Examples of the multi-image scenarios with a total of 13 tasks. The correct answers are marked in blue.}
    \label{fig: example}
    \vspace{-2mm}
\end{figure*}

In this paper, to comprehensively evaluate the multi-image ability of MLLMs, we propose a large-scale multi-image benchmark \textbf{MIBench}, which covers 13 different tasks with a total of 13K high-quality samples.
As shown in Figure \ref{fig: overview}, MIBench contains three multi-image scenarios, \textit{i.e.} \textbf{Multi-Image Instruction (MII)}, \textbf{Multimodal Knowledge-Seeking (MKS)} and \textbf{Multimodal In-Context Learning (MIC)}.
MII is a basic multi-image scenario, where the instructions involve perception, comparison and reasoning across multiple images.
MKS presents a different scenario, in which models are provided with interleaved image-text data as external knowledge, while the question itself is about a single image or even independent of any image.
MIC is another scenario where MLLMs respond to queries (\textit{e.g.} image \& question) by conditioning on a series of multimodal demos.
The three scenarios are further divided into 13 different tasks, and examples are shown in Figure \ref{fig: example}.

The MII and MKS scenarios comprise 9K multiple-choice questions.
To get these questions, we first sample images from nine existing datasets, and convert the original annotations into questions and ground truth options according to the tasks.
To obtain challenging distractors and mitigate inherent biases of options, we devise task-specific strategies to sample from annotations or generate distractors using GPT-4.
For MKS, we also devise corresponding strategies to sample images and associated texts from the datasets as external knowledge.
The MIC scenario contains 4K short-answer questions, covering close-ended VQA, open-ended VQA, object hallucination, and demo-based task learning.
We convert the data sampled from four datasets into the VQA format, and retrieve samples of the same task to construct demos.
To ensure high quality, we combine automated filtering and manual verification to remove samples with ambiguous or duplicate options.
For multiple-choice questions, we use accuracy as the metric and employ circular evaluation (\citealp{liu2023mmbench}) to mitigate the position bias of LLMs.
For short-answer questions, we use exact matching as the metric.

We evaluate several existing MLLMs on the proposed MIBench, including both closed-source (\textit{e.g.} GPT-4o) and open-source models (\textit{e.g.} LLaVA-1.5, Idefics2 and mPLUG-Owl3).
The evaluation results reveal that current MLLMs especially open-source models have major flaws in multi-image scenarios.
The annotated data of our MIBench is publicly available to spur progress in improving the multi-image abilities of MLLMs.

Our contributions are summarized as follows:

\begin{itemize}
    \item We propose the first large-scale and comprehensive benchmark MIBench for evaluating the multi-image abilities of MLLMs, covering three scenarios and 13 tasks in total.
    \item The evaluation on MIBench reveals that existing MLLMs especially open-source models face significant challenges in \textbf{fine-grained perception} and \textbf{multi-image reasoning}.
    \item Current MLLMs perform poorly in the \textbf{multimodal knowledge-seeking} scenario. And there still exists considerable room for improvement in the \textbf{multimodal in-context learning} abilities.
\end{itemize}

\section{Related Work}

\subsection{Multimodal Large Language Models}

Recent research \citep{zhu2023minigpt,liu2024visual, dai2024instructblip,ye2023mplug} has expanded LLMs (\textit{e.g.} LLaMA \citealp{touvron2023llama}) into multimodal scenarios, enabling them to process both visual and textual information. Some studies \citep{jiang2024mantis, huang2023sparkles, laurenccon2024matters, ye2024mplug3} have further explored augmenting MLLMs with multi-image understanding abilities. However, there lacks a comprehensive benchmark for evaluating the multi-image abilities of MLLMs, which limits the full exploration of these models' potential and hinders the development of this field.

\subsection{MLLM Benchmarks}

The rapid development of MLLMs has led to the emergence of a series of benchmarks, such as LVLM-eHub \citep{xu2023lvlm}, MMBench \citep{liu2023mmbench}, MM-Vet \citep{yu2023mm} and SEED-Bench \citep{li2023seed}. However, these benchmarks primarily focus on single-image evaluation, and often overlook multi-image perception and reasoning abilities, which hold even greater practical value.
Some recent studies develop benchmarks for assessing multi-image capabilities. Sparkles-Eval aims to establish a benchmark for multi-turn dialogues and multi-image scenarios. However, it exclusively focuses on the dialogue scenario, and relies entirely on GPT-4 for evaluation. Besides, it has a small data scale. Other datasets such as Mantis-Eval \citep{jiang2024mantis} and SEED-Bench2 \citep{li2024seed} also cover a small number of multi-image tasks, with a limited scale due to reliance on manual annotation.

In this paper, we propose a large-scale benchmark covering three multi-image scenarios and 13 tasks, to comprehensively evaluate the multi-image capabilities of MLLMs.

\section{MIBench}

\subsection{Evaluation Taxonomy}
\label{sec: taxonomy}

We categorize multi-image inputs into three scenarios: Multi-Image Instruction (MII), Multimodal Knowledge-Seeking (MKS) and Multimodal In-Context Learning (MIC).
As Figure \ref{fig: example} shows, MII refers to cases where instructions involve perception, comparison and reasoning across multiple images.
For instance, ``Do the two images show the same number of cats?''
MKS examines the ability of MLLMs to acquire relevant information from external knowledge, which is provided in an interleaved image-text format.
Compared to MII, the questions in the MKS scenario can be about a single image or even independent of any visual content.
MIC is another popular scenario, in which MLLMs respond to visual questions while being provided with a series of multimodal demonstrations (\textit{i.e.}, demos).

\subsubsection{Multi-Image Instruction}
According to the semantic types of the instructions, we further categorize the Multi-Image Instruction scenario into the following five tasks.

\vspace{0.3em}
\noindent\textbf{General Comparison (GC)} task examines the model's general understanding of each image (\textit{e.g.} scene, attribute and location), and comparison across different images. GC represents the most fundamental aspect of multi-image abilities.
We use the image-pair description dataset NLVR2 \citep{suhr2018corpus} for data construction.

\vspace{0.3em}
\noindent\textbf{Subtle Difference (SD)} task examines the model's ability to perceive subtle differences between similar images. Compared to general comparison, the SD task requires more fine-grained perception ability.
The image editing dataset MagicBrush \citep{zhang2024magicbrush} is adopted in this task.

\vspace{0.3em}
\noindent\textbf{Visual Referring (VR)} task evaluates whether the model can utilize the referring information provided by input images to comprehend the relationships between different objects. Figure \ref{fig: example}(c) shows an example of the VR task, whose data is from the visual relation dataset VrR-VG \citep{liang2019vrr}.

\vspace{0.3em}
\noindent\textbf{Temporal Reasoning (TR)} task assesses the model's understanding of the temporal relationships among a series of consecutive images, and its comprehension of the overall content conveyed by these images.
We employ the video understanding dataset Something-Something V2 \citep{goyal2017something} for this task.

\vspace{0.3em}
\noindent\textbf{Logical Reasoning (LR)} task requires the model to perform logical reasoning and analyze the causal relationships between objects or events shown in the input images.
The video QA dataset NExT-QA \citep{xiao2021next} is used for data construction.

\subsubsection{Multimodal Knowledge-Seeking} 
Based on the forms of external knowledge, we categorize the Multimodal Knowledge-Seeking scenario into the following four tasks.

\vspace{0.3em}
\noindent\textbf{Fine-grained Visual Recognition (FVR)} task examines the model's ability to recognize the object in the query image when given multiple reference images. It requires the model to understand the image-label correspondence in the reference images, as well as link similar images.
A combination of several fine-grained recognition datasets (\citealp{khosla2011novel}, \citealp{wah2011caltech} and \citealp{nilsback2008automated}) is used for this task.

\vspace{0.3em}
\noindent\textbf{Text-Rich Images (TRI)} VQA task evaluates the model's ability to understand text-rich images and extract information relevant to the question, which is very common in real-world scenarios (\textit{e.g.} reading slides or documents).
We adopt the SlideVQA \citep{tanaka2023slidevqa} dataset for data construction.

\vspace{0.3em}
\noindent\textbf{Vision-linked Textual Knowledge (VTK)} task corresponds to a very practical scenario where the question is beyond the visual content of the query image, such as querying background knowledge.
The provided external knowledge encompasses images and corresponding text which are possibly retrieved from a knowledge base (\textit{e.g.}, Wikipedia). The model is required to link the query image to the relevant image, and extract useful information from the corresponding text. Figure \ref{fig: example}(h) shows an example, whose data is from the InfoSeek dataset \citep{chen2023can}.

\vspace{0.3em}
\noindent\textbf{Text-linked Visual Knowledge (TVK)} task refers to cases where the text-only question is about the visual attributes of a specific object. For instance, "Is the China National Stadium round or square?"
When provided with external knowledge in an interleaved image-text form, the model needs to link the question to the relevant text, and extract visual information from the corresponding image. This task is very common in real life such as browsing web pages. Figure \ref{fig: example}(i) shows an example, whose data is from the WebQA dataset \citep{chang2022webqa}.

\subsubsection{Multimodal In-Context Learning}
The in-context learning ability enables LLMs to gain performance boost when provided with a series of demos.
Recent studies \citep{alayrac2022flamingo,awadalla2023openflamingo,laurenccon2024matters} have also explored multimodal in-context learning (MIC).
For the evaluation of the MIC ability, existing methods solely assess the model's performance via a holistic metric, such as accuracy on the VQAv2 \citep{goyal2017making} dataset.
To evaluate the model's MIC ability in a fine-grained manner, we categorize the MIC scenario into the following four distinct tasks.

\vspace{0.3em}
\noindent\textbf{Close-ended VQA} task requires the model to answer from a predefined set of responses, which is provided via multimodal demos.
This task examines the model's ability to learn the image-label mapping relationships from the demos.
We use the Mini-ImageNet dataset \citep{vinyals2016matching} for data construction.

\vspace{0.3em}
\noindent\textbf{Open-ended VQA} task has an open range of possible answers which cannot be fully covered by the provided demos. The task evaluates the model's ability to learn task patterns through demos.
We conduct a balanced sampling of different knowledge types from the OK-VQA dataset \citep{marino2019ok} for this task.

\vspace{0.3em}
\noindent\textbf{Hallucination} phenomenon is a significant challenge faced by MLLMs. In this task, we convert the hallucination dataset POPE \citep{li2023evaluating} into in-context learning format, and study the impact of the model's MIC ability on the hallucination phenomenon.

\vspace{0.3em}
\noindent\textbf{Demo-based Task Learning} is a core aspect of in-context learning, which enables the model to rapidly adapt to new tasks given a few demos.
To investigate existing MLLMs' demo-based task learning ability, we select several visual tasks from the VQAv2 dataset and remove the task instructions. Instead, we present the task demos in the form like ``rabbit: 3''. Figure \ref{fig: example}(m) shows an example.

\subsection{Data Generation}
\label{sec: data_generation}

In Section \ref{sec: taxonomy}, we introduced the evaluation tasks and the corresponding data source of the proposed MIBench.
However, the generation of test samples using the original datasets is nontrivial.
We meticulously devise a data generation pipeline, including various strategies of question generation, distractor generation and external knowledge sampling for different tasks.

\vspace{0.3em}
\noindent\textbf{Question Generation.} To enhance the diversity of questions, we devise corresponding prompts for the tasks, and employ GPT-4 to generate a variety of question forms. We then randomly sample from the question pool to construct the test samples. For instance, for the General Comparison (GC) task, the questions such as ``Is the subsequent sentence an accurate portrayal of the two images?'' and ``Can the given sentence accurately illustrate what's in these two images?'' are utilized.

\vspace{0.3em}
\noindent\textbf{Distractor Generation.} For different tasks, we adopt two methods of distractor generation. One way is to sample from the original annotations following certain strategies. For instance, for the Temporal Reasoning (TR) task, we utilize the Something-something V2 dataset for data construction. To prevent the model from taking shortcuts by identifying objects to choose the correct options, we sample different temporal relationships of the same object from the annotations as distractors. In this way, the constructed test samples can more accurately reflect the model's understanding of temporal relationships.
The second method is to generate distractors with the help of GPT-4. For instance, in the Text-Rich Images (TRI) VQA task, we prompt GPT-4 to generate distractors according to the question and the correct answer.

\vspace{0.3em}
\noindent\textbf{External Knowledge Sampling.} For the Multimodal Knowledge-Seeking (MKS) scenario, reasonably sampling interleaved image-text data as external knowledge is very important to the quality of test samples. For instance, in the Vision-linked Textual Knowledge (VTK) task, we select text and images from the original annotations which have the same question as the current query but with different answers as external knowledge. This approach avoids selecting text and images unrelated to the current query, and can thus generate more challenging distractors.
Additionally, some datasets require more complex information extraction. For instance, we use GPT-4 to extract question-related segments from the original wiki entries of the InfoSeek dataset, which can be as long as several thousand words.

\subsection{Quality Control}

To mitigate data contamination, our construction of test data exclusively utilizes the validation or test sets from existing datasets.
Furthermore, we combine automated filtering and manual verification to ensure the quality and reliability of the test data.

Specifically, after the data generation process, we perform two automated filtering strategies on the obtained data.
\textbf{1)} We remove images from the input samples, and test multiple advanced MLLMs on them. Then we discard samples which can still be answered correctly without visual input. This avoids the overestimation of model performance due to the textual bias of the questions and options.
\textbf{2)} For the Multimodal Knowledge-Seeking scenario, we eliminate external knowledge from the samples and test them using multiple MLLMs. Then we remove samples which the models can answer correctly without external knowledge. This mitigates the impact of internal knowledge of the model, and provides a more accurate assessment of the model's ability of utilizing external knowledge.

As stated in Section \ref{sec: data_generation}, for some tasks such as Visual Referring, we employ GPT-4 to generate distractors. To ensure the high quality of the generated samples, we apply manual verification after automated filtering. The process is conducted by three trained annotators who possess relevant professional backgrounds. Specifically, a sample is discarded if there are duplicate options or more than one correct option.

\subsection{Evaluation}
For the multiple-choice questions, we employ the accuracy of the predicted options as the evaluation metric. Notably, early MLLMs such as mPLUG-Owl tend to produce longer responses rather than directly outputting the option. For these models, we use GPT-4 to determine which option matches the predicted content. 
In addition, similar to the observation of MMBench, we find that different MLLMs show preferences for specific options (\textit{i.e.} position bias).
Therefore, we set the correct option sequentially to ``A'', ``B'', ``C'' and ``D''. A model is considered to have correctly answered a sample only if it consistently provides the correct response across multiple tests. 
In this way, the impact of position bias on the evaluation results is mitigated.

\section{Experiments}

\begin{table*}[t]
\centering
\small
\resizebox{0.95\textwidth}{!}{
\begin{tabular}{l|ccccc|cccc}
\toprule
\multirow{2}{*}{\textbf{Model}} & \multicolumn{5}{c|}{\textbf{Multi-Image Instruction}} & \multicolumn{4}{c}{\textbf{Multimodal Knowledge-Seeking}} \\
 & \textbf{GC} &
\textbf{SD} &
\textbf{VR} &
\textbf{TR} &
\textbf{LR} &
\textbf{FVR} &
\textbf{TRI} &
\textbf{VTK} &
\textbf{TVK} \\
\midrule

\rowcolor{gray!20}\multicolumn{10}{c}{\textbf{Closed-source MLLMs}} \\
\midrule
GPT-4o & 80.7 & 90.5 & 46.8 & 68.0 & 69.8  & 98.3 & 74.8 & 54.7 & 63.3  \\
GPT-4V & 72.8 & 79.2 & 45.8 & 61.8 & 66.3  & 90.2 & 71.0 & 52.0 & 56.0   \\
\midrule

\rowcolor{gray!20}\multicolumn{10}{c}{\textbf{Open-source MLLMs}} \\

\midrule
mPLUG-Owl3 & 86.4 & 70.1 & 33.0 & 46.8 & 67.2 & 76.4 & 50.1 & 31.1 & 48.8 \\
Mantis & 83.0 & 54.1  & 37.6 & 45.5 & 63.4  & 16.4 & 37.7 & 26.4 & 41.7 \\
Idefics2-I & 83.1 & 49.7  & 32.6 & 44.8 & 56.4 & 42.4 & 43.9 & 25.6 & 39.0 \\
MMICL & 53.7 & 46.4 & 41.1 & 47.0 & 59.6 & 56.6 & 27.6 & 22.1 & 35.9 \\

mPLUG-Owl2 & 64.2 & 40.1 & 35.6 & 30.7 & 41.3 & 13.3 & 39.0 & 17.0  & 25.6 \\
Qwen-VL & 45.9 & 22.5 & 16.3 & 27.5 & 36.8 & 58.8 & 35.9 & 22.9 & 18.1 \\
LLaVA-1.5 & 40.6 & 14.9 & 24.1 & 30.1 & 44.8 & 18.2 & 25.8 & 16.7 & 26.3 \\

mPLUG-Owl & 19.1 & 4.0 & 21.7 & 8.0 & 29.2 & 17.3 & 12.1 & 14.9 & 20.6 \\

\bottomrule
\end{tabular}}
\caption{Evaluation results on the multi-image instruction and multimodal knowledge-seeking scenarios of MIBench.}
\label{tab:results}
\end{table*}

\begin{figure}[t]
    \centering
    \includegraphics[width=\linewidth]{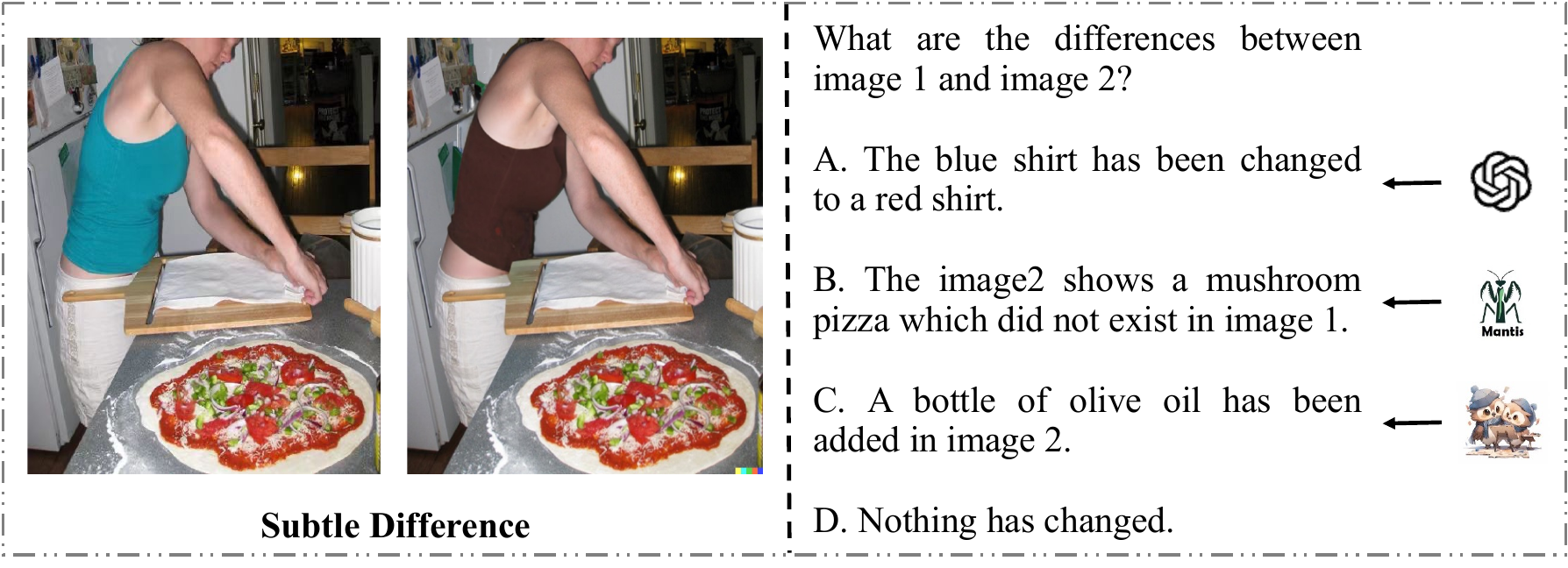}
    \caption{A qualitative case of the Subtle Difference task, where open-source MLLMs show inferior performance due to limited fine-grained perception ability.}
    \label{new_fig}
    \vspace{-2mm}
\end{figure}

\subsection{Models}

In this section, we evaluate MLLMs using the constructed MIBench dataset. 
We first evaluate MLLMs on the Multi-Image Instruction and Multimodal Knowledge-Seeking scenarios. These models can be categorized into three distinct groups: (1) closed-source models, including GPT-4V and GPT-4o; (2) open-source single-image MLLMs, including mPLUG-Owl \citep{ye2023mplug}, LLaVA-1.5 \citep{liu2024visual}, Qwen-VL \citep{bai2023qwen} and mPLUG-Owl2 \citep{ye2024mplug}; (3) open-source models natively supporting multi-image input, including Mantis \citep{jiang2024mantis}, Idefics2 \citep{laurenccon2024matters} and mPLUG-Owl3 \citep{ye2024mplug3}.
For the open-source models, we employ greedy decoding for prediction generation. 

Then we evaluate open-source MLLMs on the Multimodal In-Context Learning (MIC) scenario.
However, as most of these models have neither been pre-trained on large-scale interleaved image-text data nor fine-tuned on ICL format data, they do not exhibit MIC capabilities.
In the tests across the four MIC tasks, they consistently exhibit a negative ICL effect, \textit{i.e.}, their performance decreases as the number of demos increases.
Therefore, we only present the evaluation results of models that possess multimodal ICL abilities, \textit{i.e.} OpenFlamingo, MMICL, IDEFICS and IDEFICS-I.

\subsection{Evaluation Results}

\begin{figure*}
    \centering
    \includegraphics[width=\linewidth]{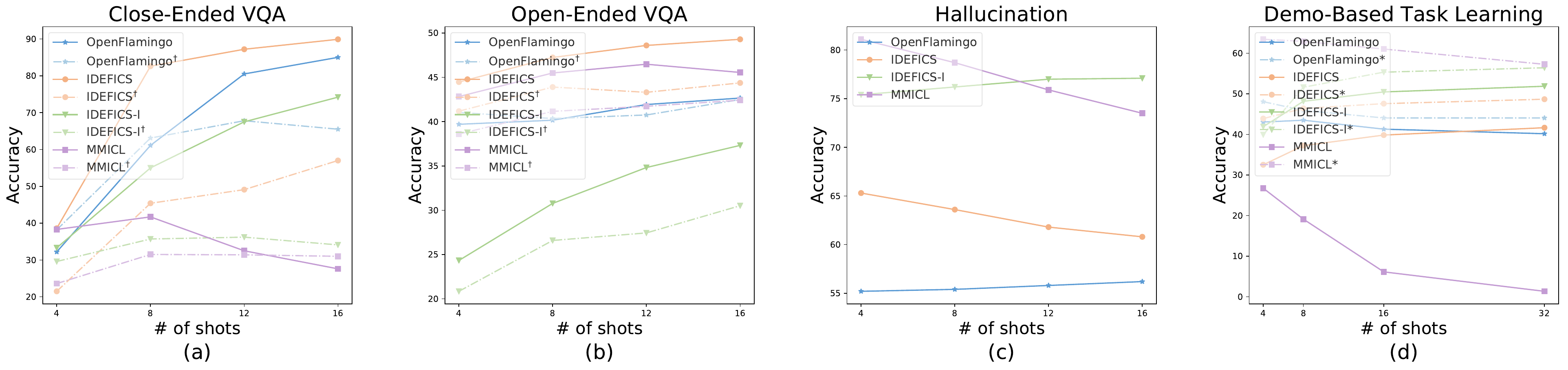}
    \caption{Evaluation results on the Multimodal In-Context Learning scenario.}
    \label{fig: exp_mic}
\end{figure*}

\subsubsection{Multi-Image Instruction \& Multimodal Knowledge-Seeking}

Table \ref{tab:results} shows the main results of the Multi-Image Instruction (MII) and Multimodal Knowledge-Seeking (MKS) scenarios. Through these results, we have several valuable observations:

\vspace{0.3em}
\noindent\textbf{Closed-source MLLMs exhibit superior performance than open-source models.} In most MII and MKS tasks, the performance of open-source models lags significantly behind that of proprietary models.
For instance, on the Temporal Reasoning (TR) task, the best-performing open-source model MMICL achieves an accuracy of 47.0\%, falling behind GPT-4o by 21.0\%.

\vspace{0.3em}
\noindent\textbf{Open-source MLLMs are inadequate in fine-grained perception tasks.}
Although many open-source MLLMs have decent performance on the General Comparison (GC) task, their performance is significantly worse on the Subtle Difference (SD) and Text-Rich Images (TRI) VQA tasks.
For instance, Idefics2-I achieves 83.1\%, 49.7\% and 43.9\% on the three tasks respectively.
In contrast, GPT-4V and GPT-4o largely outperform open-source models, due to their high-resolution input strategy.
Figure~\ref{new_fig} provides a qualitative case supporting the above point.

\vspace{0.3em}
\noindent\textbf{Multi-image MLLMs perform better than single-image models on most tasks.}
This verifies that pre-training on interleaved image-text data (\textit{e.g.} Idefics2-I) and instruction tuning on multi-image data (\textit{e.g.} Mantis) are both beneficial for improving multi-image abilities.
Combining multi-image pre-training and instruction tuning, mPLUG-Owl3 achieves better performance than other open-source MLLMs on most tasks.

\vspace{0.3em}
\noindent\textbf{The Visual Referring (VR) task is particularly challenging for existing MLLMs.} As it requires integration of fine-grained perception, spatial correspondence and relation reasoning, most models have not achieved satisfactory performance on the VR task. Even the best-performing model GPT-4o has not exceeded a 50\% accuracy rate.

\vspace{0.3em}
\noindent\textbf{Most existing open-source MLLMs perform poorly in the Multimodal Knowledge-Seeking (MKS) scenario.}
For instance, the accuracy rates of mPLUG-Owl2 on both the Vision-linked Textual Knowledge (VTK) and Text-linked Visual Knowledge (TVK) tasks are below 30\%. In the Fine-grained Visual Recognition (FVR) task, which requires the combination of fine-grained perception and comparison abilities, mPLUG-Owl2's performance is merely over 10\%. Compared to single-image MLLMs, multi-image models such as Idefics2-I exhibit better capabilities in utilizing multimodal external knowledge. 
However, there is still significant room for improvement, as the performance of Idefics2-I on both the VTK and TVK tasks is under 40\%.

\begin{table*}[t]
\centering
\footnotesize{
\resizebox{0.99\textwidth}{!}{
\begin{tabular}{l|cc|cc|cc}
\toprule
& \multicolumn{2}{c|}{\textbf{Text-rich Images VQA}} & \multicolumn{2}{c|}{\textbf{Text-linked Visual Knowledge}} & \multicolumn{2}{c}{\textbf{Vision-linked Textual Knowledge}} \\
                             & \text{w/ Dis.}       & \text{w/o Dis.}      & \text{w/ Dis.}       & \text{w/o Dis.}     & \text{w/ Dis.}        & \text{w/o Dis.}       \\ \midrule
mPLUG-Owl2          &            39.0               &         42.1                  & 25.6                     & 29.6                  &        17.0                    &             90.1                \\
Mantis                & 37.7                  & 42.6                  & 41.7                     & 47.7                  & 26.4                      & 88.1                       \\
Idefics2-${\mathrm{I}}$             & 43.9                     & 46.8 (59.5)             & 39.0                     & 45.2                  & 25.6                      & 91.0                       \\ 
\bottomrule
\end{tabular}}}
\caption{Ablation study of the impact of distractors on various tasks on the multimodal knowledge-seeking scenario.}
\label{tab:my_table_knowledge}
\vspace{-2mm}
\end{table*}

\begin{table}[t]
\centering
\small
\resizebox{0.99\linewidth}{!}{
\begin{tabular}{l|cc|cc}
 \toprule
 & \multicolumn{2}{c|}{\textbf{Confusion}} & \multicolumn{2}{c}{\textbf{Reasoning}}\\
  & \text{Conf. A} & \text{Conf. B} & Tem. & Obj.  \\
\midrule
 mPLUG--Owl2 & 87.0 & 25.0 & 30.7 & 56.6 \\
 Qwen--VL & 89.2 & 26.8 & 27.5 & 60.9 \\
 LLaVA--1.5 & 91.8 & 31.6 & 30.1 & 59.3 \\
 Mantis & 91.2 & 83.6 & 45.5 & 75.7 \\
\bottomrule
\end{tabular}
}
\caption{Ablation study on the multi-image confusion phenomenon and the temporal reasoning task.}
\label{tab:my_table_confusion}
\vspace{-2mm}
\end{table}

\subsubsection{Multimodal In-Context Learning}
Figure \ref{fig: exp_mic} shows the performances of OpenFlamingo, MMICL, and IDEFICS on multimodal ICL scenarios. The horizontal axis represents different shots (\textit{i.e.}, the number of demos), and the vertical axis represents accuracy. 
To investigate the impact of images on multimodal ICL, the models that remove the images from demos ($\dag$ in Figure \ref{fig: exp_mic}) are evaluated on close-ended VQA and open-ended VQA.

\vspace{0.3em}
\noindent\textbf{The current models exhibit multimodal ICL abilities on close-ended VQA.} As shown in Figure \ref{fig: exp_mic}(a), 
after removing the images in the demos, the performance of most models declines, and the extent of this decline increases with the number of shots. This indicates that these models have learned the image-label mapping relationships in the demos, exhibiting a certain degree of multimodal ICL ability.

\vspace{0.3em}
\noindent\textbf{Multimodal ICL abilities of different models appears to be driven by different modalities.}
As shown in Figure \ref{fig: exp_mic}(b), when the number of demos increases, all models show consistent performance improvement. However, for OpenFlamingo, removing images from the demos does not cause a significant performance change, indicating that OpenFlamingo's ICL on this task is primarily driven by text. In contrast, the absence of images leads to a significant performance decline for IDEFICS and MMICL, indicating that they possess a certain degree of multimodal ICL ability.

\vspace{0.3em}
\noindent\textbf{Multimodal ICL abilities of current models do not alleviate the hallucination phenomenon.} As shown in Figure \ref{fig: exp_mic}(c), on object hallucination task, only IDEFICS-I and Idefics2-I exhibit slight accuracy improvements with an increasing number of shots, while other models show negative effects. It indicates that ICL provides very limited help in mitigating hallucinations and may even exacerbate them. Comparing the base and instruction-tuned versions of IDEFICS, it is evident that instruction tuning can help alleviate hallucinations.

\vspace{0.3em}
\noindent\textbf{Most models possess some capacity of demo-based task learning, but the capacity is relatively limited.} 
Figure \ref{fig: exp_mic}(d) shows the model performance under different shots in counting and color tasks demonstrated only through examples. It is evident that with an increasing number of demos, IDEFICS shows significant gains, OpenFlamingo quickly reaches saturation, and MMICL even fails to follow the task format presented in the demos. In fact, except for MMICL, these models can effectively follow the output format with just 4 shots, and their performance improves with more shots. It reflects that OpenFlamingo and IDEFICS possess a certain degree of demo-based task learning ability. 
In addition, compared to the experimental results with explicit task instructions (\textit{e.g.}, `How many people are in the room?'), there remains a significant performance gap, indicating that the demo-based task learning abilities of current models still have substantial room for improvement.

\vspace{-2mm}
\subsection{Analysis}

\subsubsection{Multi-image Confusion Phenomenon}

When evaluating MLLMs on the MIBench benchmark, we observe that open-source MLLMs, particularly single-image models, exhibit confusion when handling multiple images. To validate this issue, we derive two confusion subsets with 500 samples respectively from the POPE dataset used in the hallucination task. In subset A, each sample consists of one image and one question. The question asks whether a specific object is present in the image, which actually is not contained. In subset B, an extra image containing the object in the question is added to each sample in subset A as a distractor. As shown in Table \ref{tab:my_table_confusion}, it can be observed that the performance of the three single-image models significantly decline after the addition of the extra image, while the multi-image model Mantis also has a slight performance drop. It confirms that current open-source MLLMs, especially single-image models, suffer from severe confusion, thereby affecting their performance in multi-image instructions and multimodal knowledge-seeking.

\subsubsection{Limited Reasoning Ability}

In the construction of temporal reasoning, we utilize the ground truth of videos as the correct option and sample different actions of the same object as distractors. Under this setting, the majority of MLLMs demonstrate poor performance. To further study these results, we replace the same objects in the distractors with different objects and test several representative models. As indicated in the table, under the setting where distractors contain different objects, the performance of mPLUG-Owl2, LLaVA-1.5 and Mantis models significantly improves, as the models can take shortcuts by distinguishing between objects. The results indicate that for current MLLMs, the reasoning ability across multiple images is significantly inferior to their spatial perception ability.

\subsubsection{Bottlenecks of the MKS task}
\label{sec: analysis mks}
Compared to multi-image instruction, multimodal knowledge-seeking requires the model to extract relevant information from external image-text knowledge sources and is thus more challenging.
To investigate the bottlenecks of multimodal knowledge-seeking tasks,
we compare the impact of distracting content.

As shown in Table \ref{tab:my_table_knowledge}, for text-linked visual knowledge, removing distracting content(\textit{i.e.}, only retaining the information relevant to the question) results in a certain performance improvement. It indicates that the model's ability to identify relevant information (\textit{i.e.}, link by text) still can be improved. On the other hand, even after the removal of distracting content, the performance remains poor. It suggests that the primary bottleneck for this task is the deficiencies of MLLMs in perceiving and reasoning with visual information. 

In contrast, for the task of vision-linked textual knowledge, the removal of distracting content leads to a significant performance improvement. It suggests that the bottleneck for this task lies in the MLLMs' ability to mine effective messages through image comparison (\textit{i.e.}, link by image). 

On text-rich images VQA, removing distracting content brings some performance boost. Based on this, Idefics2-I further boosts from 46.8\% to 59.5\% by employing image splitting for higher resolution inputs. 
The significant performance gain indicates that the bottleneck of this task is more related to information loss caused by low resolution. 

From the above comparisons, it can be concluded that the current MLLMs' abilities in perceiving, contrasting, and reasoning with visual information are remarkably inferior to their abilities with text, and face substantial challenges in understanding rich-text images due to resolution issues.

\section{Conclusion}
While MLLMs have shown strong performance in various vision-language tasks, their abilities with multi-image inputs remain underexplored. To address this, we introduce MIBench in this paper, a benchmark that evaluates MLLMs across three multi-image scenarios: multi-image instruction, multimodal knowledge-seeking and multimodal in-context learning, covering 13 tasks with 13K annotated samples. The evaluation results reveal that existing models, despite excelling in single-image tasks, face significant challenges with multi-image inputs.
The annotated data is publicly available to facilitate further research. We hope this work can spur progress in improving the multi-image abilities of MLLMs.

\section*{Limitations}
Due to the input length limitation of current MLLMs, the Multi-Image Instruction and Multimodal Knowledge-Seeking scenarios of our benchmark include 2 to 8 input images in each sample. However, real-world scenarios may involve a large number of images.
We'll investigate the evaluation of MLLMs over more images in future work.

\section*{Acknowledgement}
This work is supported by Beijing Natural Science Foundation (JQ21017, L243015, L223003), the National Key Research and Development Program of China (No. 2020AAA0105802), the Natural Science Foundation of China (No. 62036011, 62192782), and the Project of Beijing Science and Technology Committee (No. Z231100005923046).

\bibliography{custom}

\clearpage

\begin{table*}[t]
\centering
\resizebox{0.98\textwidth}{!}{
\begin{tabular}{lccc}
\toprule
 Task & Image Source & Question Source & Distractor Source \\
 \midrule
 General Comparison & NLVR2 & GPT-4 generated & Original annotations \\
 Subtle Difference & MagicBrush & GPT-4 generated & Sampled from annotations \\
 Visual Referring & VrR-VG & Manual & GPT-4 generated \\
 Temporal Reasoning & Something-Something V2 & Manual & Sampled from annotations \\
 Logical Reasoning & NeXT-QA & Original annotations & Original annotations \\
 Fine-grained Visual Recognition & Dogs / Birds / Flowers / Cars & GPT-4 generated & Sampled from annotations \\
 Text-Rich Images & SlideVQA & Original annotations & GPT-4 generated \\
 Vision-linked Textual Knowledge & InfoSeek & Extracted from annotations & Sampled from annotations \\
 Text-linked Visual Knowledge & WebQA & Sampled from annotations & GPT-4 generated \\
 Close-ended VQA & Mini-ImageNet & Sampled from annotations & - \\
 Open-ended VQA & OKVQA & Sampled from annotations & - \\
 Hallucination & POPE & Sampled from annotations & - \\
 Demo-based Task Learning & VQAv2 & Converted from annotations & - \\
\bottomrule
\end{tabular}}
\caption{More details of the data generation process.}
\label{tab:task_sources}
\end{table*}

\begin{table*}[t]
\centering
\resizebox{0.98\textwidth}{!}{
\begin{tabular}{lccc}
\toprule
Task & Image Number Per Sample & Average Question Length & Average Answer Length \\
\midrule
General Comparison & 2 & 33.3 & 1.0 \\
Subtle Difference & 2 & 19.0 & 10.0 \\
Visual Referring & 3 & 27.0 & 6.9 \\
Temporal Reasoning & 8 & 39.0 & 6.2 \\
Logical Reasoning & 8 & 44.7 & 3.1 \\
Fine-grained Visual Recognition & 5 & 35.4 & 2.6 \\
Text-Rich Images & 4 & 25.9 & 2.9 \\
Vision-linked Textual Knowledge & 5 & 562.7 & 1.7 \\
Text-linked Visual Knowledge & 4 & 76.7 & 3.6 \\
Close-ended VQA & 5-17 & 5.0 & 1.4 \\
Open-ended VQA & 5-17 & 8.1 & 1.2 \\
Hallucination & 5-17 & 7.2 & 1.0 \\
Demo-based Task Learning & 5-33 & 3.2 & 1.1 \\
\midrule
Overall & 125K (in total) & 68.2 & 4.1 \\
\bottomrule
\end{tabular}}
\caption{Image number, average question/answer length of each task.}
\label{tab:statistics}
\end{table*}

\appendix
\section{More Details of MIBench}

Table \ref{tab:task_sources} presents the detailed data generation information of each task.
Note that ``sampled from annotations'' isn't simple random sampling from the original annotations. Instead, as stated in Section \ref{sec: data_generation}, it involves designing specific sampling strategies tailored to the task.

Table \ref{tab:statistics} shows the detailed statistics of each task, including image number per sample, average question length and average answer length.
Note that ``Image Number Per Sample'' for the Multimodal In-Context (MIC) learning scenario is a range determined by the number of demos. And the whole benchmark has 125K images in total. ``Average Answer Length'' refers to the average length of options for multiple-choice questions and the average length of answers for short-answer questions.



\end{document}